\documentclass[journal,transmag]{IEEEtran}
\ifCLASSINFOpdf
\else
\fi

\usepackage[T1]{fontenc}
\usepackage{tabularray}
\usepackage{graphicx}
\usepackage{longtable}
\usepackage[table]{xcolor}
\usepackage{array}
\usepackage{color,soul}
\usepackage{pifont}
\let\oldding\ding
\renewcommand{\ding}[2][1]{\scalebox{#1}{\oldding{#2}}}
\newcolumntype{L}[1]{>{\raggedright\let\newline\\\arraybackslash\hspace{0pt}}m{#1}}
\newcolumntype{C}[1]{>{\centering\let\newline\\\arraybackslash\hspace{0pt}}m{#1}}
\usepackage{pdflscape} 
\DeclareUnicodeCharacter{2212}{-}
\usepackage{url}
\usepackage{multirow}
\usepackage{cite}
\usepackage{float}

\begin{document}
%
\title{TinyML: Tools, Applications, Challenges, and Future Research Directions}


\author{\IEEEauthorblockN{
Rakhee Kallimani\IEEEauthorrefmark{1},
Krishna Pai\IEEEauthorrefmark{2},
Prasoon Raghuwanshi\IEEEauthorrefmark{3}
Sridhar Iyer\IEEEauthorrefmark{4}, and
Onel L. A. López\IEEEauthorrefmark{5}}

\IEEEauthorblockA{\IEEEauthorrefmark{1}Department of Electrical and Electronics Engineering,\\
KLE Technological University Dr MSSCET, Belagavi, Karnataka, India - 590008.\\
Email: rakhee.kallimani@klescet.ac.in 
}
\IEEEauthorblockA{\IEEEauthorrefmark{2}Department of Electronics and Communication Engineering, \\
KLE Technological University Dr. MSSCET, Belagavi, Karnataka, India - 590008.\\
Email: krishnapai271999@gmail.com 
}
\IEEEauthorblockA{\IEEEauthorrefmark{3}Faculty of Information Technology and Electrical Engineering, University of Oulu, Finland- 90014. \\
Email: Prasoon.Raghuwanshi@oulu.fi 
}
\IEEEauthorblockA{\IEEEauthorrefmark{4}Department of Artificial Intelligence, \\
KLE Technological University Dr. MSSCET, Belagavi, Karnataka, India - 590008.\\
Email: sridhariyer1983@klescet.ac.in
}
\IEEEauthorblockA{\IEEEauthorrefmark{5}Faculty of Information Technology and Electrical Engineering, University of Oulu, Finland- 90014. \\
Email: onel.alcarazlopez@oulu.fi
}

\thanks{
Corresponding author: K. Pai (email: krishnapai271999@gmail.com).}}

%




\IEEEtitleabstractindextext{%
\begin{abstract}
In recent years, Artificial Intelligence (AI) and Machine learning (ML) have gained significant interest from both, industry and academia. Notably, conventional ML techniques require enormous amounts of power to meet the desired accuracy, which has limited their use mainly to high-capability devices such as network nodes. However, with many advancements in technologies such as the Internet of Things (IoT) and edge computing, it is desirable to incorporate ML techniques into resource-constrained embedded devices for distributed and ubiquitous intelligence. This has motivated the emergence of the TinyML paradigm which is an embedded ML technique that enables ML applications on multiple cheap, resource- and power-constrained devices. 
However, during this transition towards appropriate implementation of the TinyML technology, multiple challenges such as processing capacity optimisation, improved reliability, and maintenance of learning models’ accuracy require timely solutions. 
In this article, various avenues available for TinyML implementation are reviewed. Firstly, a background of TinyML is provided, followed by detailed discussions on various tools supporting TinyML. 
Then, state-of-art applications of TinyML using advanced technologies are detailed.
Lastly, various research challenges and future directions are identified.
\end{abstract}

\begin{IEEEkeywords}
TinyML, embedded AI, edge computing, IoT.
\end{IEEEkeywords}}

\maketitle

\IEEEdisplaynontitleabstractindextext

%
\IEEEpeerreviewmaketitle

\section{Introduction}

\IEEEPARstart{T}{he} Internet of Things (IoT) leverages edge computing to enable the seamless processing of data from millions of interconnected sensors and other devices. The IoT devices are deployable at the network edge and incur a very low memory footprint and processing capacity \cite{Goudarzi_Palaniswami_Buyya_2021, Muhammad_Hossain_2021}. The IoT ecosystems increasingly depend on the edge platforms to collect and transmit the data 
\cite{Li_Deng_She_Zhang_Wang_Ma_2021}. In fact, within the IoT ecosystem, edge devices gather sensor data, which is then transmitted to a remote cloud or a nearby location for processing \cite{Liu_Liu_Wang_Gao_Wang_2022}. The edge computing technology performs computations and stores original data, simultaneously providing the infrastructure to support distributed computing \cite{Muniswamaiah_Agerwala_Tappert_2021, Ying_Hsieh_Hou_Hou_Liu_Zhang_Wang_Pan_2021}.  Additionally, edge computing provides (i) effective privacy, security, and reliability to end-users, (ii) lower delay, (iii) higher throughput, and availability and effective response to services and applications \cite{Wu_Huang_Xie_Nie_Bao_Qin_2021, Bao_Wu_Guleng_Zhang_Yau_Ji_2021}. Notably, by using a collaborative technique among the sensors, the edge devices, and the cloud, data processing may be conducted (at least partially) at the network edge rather than at the cloud. This may facilitate quality data management, effective service delivery, data persistence, and content caching \cite{Singh_Bello_Hussein_Erbad_Mohamed_2021}. Further, for implementation in various applications such as human-to-machine (H2M) interaction and smart healthcare, edge computing provides an opportunity to significantly improve the network services which are automated simultaneously ensuring that the network back-haul is less burdened \cite{Ding_Zhou_Ma_Zhang_Hsu_Wang_2021,Mahmood.2020}.

Recent research on IoT edge computing is in the spotlight as it facilitates the implementation of Machine Learning  (ML) techniques in many use cases. However, hardware to be deployed at the network edge is severely constrained in resources such as memory capacity, power consumption, and compatibility, limiting the provisioning of high-end complex services \cite{Mahmood.2020}. In fact, the current IoT edge scenario is undermined as it does not exactly depict the envisioned cloud-to-embedded paradigm \cite{ray_2022}. In effect, edge computing, though expected to do so in the future, does not yet offer significant power saving and high transmission capacity \cite{Guleria_Das_Sahu_2021}. 
This is mainly due to the existing differences between hardware and software technologies, which lead to heterogeneous systems. Therefore, holistic and harmonious infrastructures are required, especially for training, updating, and deploying the ML models \cite{Ogino_2021, Ren_Sun_Luo_Guizani_2022}. Also, the architectures designed for embedded systems depend on the type of hardware and software, which in turn represents a hindrance to developing a standard ML architecture for all edge IoT networks. Additionally, majority existing processors which are embedded permit only processing of sensor data in a generalized manner and applications of the software. 

Currently, the large amount of data generated by multiple devices is sent to the cloud for processing due to the computationally intensive nature of existing network implementations. Indeed, operating advanced ML models such as deep neural networks (DNNs) and deep learning (DL) demand graphics processing units (GPUs) and dedicated hardware application specific integrated circuits (ASICs), which require large  energy amounts and capacity of memory. Hence, there is currently significant interest in optimizing ML algorithms to make them more energy-efficient. Concurrently, there is also a growing demand to miniaturize low power embedded devices. These aspects have paved the way for the introduction of Tiny Machine Learning (TinyML), which implements ML algorithms on tiny devices such as edge IoT devices. TinyML enables signal processing at these devices while provisioning embedded intelligence, thus constituting a paradigm shift from cloud intelligence. Indeed, TinyML is a rapidly evolving edge computing concept which links ML and embedded systems \cite{Johnny_Knutsson_Arm_2021, TinyMLHome, Warden_Situnayake}. All in all, TinyML may enable ultra low power and cost systems demonstrating efficiency and privacy \cite{Alajlan_Ibrahim_2022}. Further, in cases of inadequate connectivity, TinyML may provide on-premise analytic(s), undoubtedly appealing for IoT services.

\begin{table*}[ht!]
 \caption{Summary of Recent Surveys related to TinyML.} 
     \label{tab:summary_of_surveys}
    \centering
    \resizebox{18cm}{!}{
    \begin{tabular}{|p{1.5cm}|p{13cm}|p{4cm}|}
\hline
  \textbf{Reference} & \textbf{Key Contribution} & \textbf{Limitations w.r.t our survey} \\
  \hline 

\cite{ray_2022}& 
    \ding[0.5]{108}\quad Intuitive review regarding the possibilities for TinyML. 
    
    \ding[0.5]{108}\quad Background and toolsets to support TinyML.
    
    \ding[0.5]{108}\quad Key enablers to improve TinyML systems and state-of-art architectures for TinyML. 
    
    \ding[0.5]{108}\quad Key challenges and future road-map to mitigate numerous research issues of TinyML. &
    \ding[0.5]{108}\quad Recent case studies of TinyML.

    \ding[0.5]{108}\quad Detailed directions to highlight key research findings.    
 \\ \hline

 \cite{Alajlan_Ibrahim_2022}& 
 
     \ding[0.5]{108}\quad  Overview and review of TinyML studies.

    \ding[0.5]{108}\quad   Analysis of ML models types used for TinyML.

    \ding[0.5]{108}\quad  Details of datasets and devices types and characteristics.

    \ding[0.5]{108}\quad  Available resource constraints such as hardware platforms, and supporting platforms.
&
    \ding[0.5]{108}\quad  Specific application.    
 \\ \hline

\cite{Dutta_Bharali_2021}& 
    
    \ding[0.5]{108}\quad Definition of TinyML. 
    
    \ding[0.5]{108}\quad Background information on various related technologies.
    
    \ding[0.5]{108}\quad TinyML as a service. 
    
    \ding[0.5]{108}\quad Role of 5G for TinyML IoT.
    
    \ding[0.5]{108}\quad Recent progress in TinyML research. 
    
    \ding[0.5]{108}\quad Future oppportunities and challenges.
&
    \ding[0.5]{108}\quad Need for standardization.  
 \\ \hline
 
 \cite{Shafique_Theocharides_Reddy_Murmann_2021}& 
    
    \ding[0.5]{108}\quad Cross-layer design flow as a key aim of TinyML.

    \ding[0.5]{108}\quad TinyML applications, frameworks and benchmarking. 

    \ding[0.5]{108}\quad TinyML as a service by implementing it via  efficient hardware and software design. 
    
    \ding[0.5]{108}\quad Challenges, opportunities, and vision for the road ahead on TinyML. &
    \ding[0.5]{108}\quad Supporting platforms and available library/framework for specific applications. \\ \hline

 \cite{Immonen_Hamalainen_2022}& 
 
    \ding[0.5]{108}\quad  Resource optimization challenges of TinyML.

    \ding[0.5]{108}\quad  Present state of TinyML frameworks, libraries, development environments, and tools. 

    \ding[0.5]{108}\quad   Benchmarking of TinyML devices. 

    \ding[0.5]{108}\quad   Emerging techniques and approaches to boost/expand TinyML process, and improve data privacy and security.

    \ding[0.5]{108}\quad Future development of TinyML. 
&
    \ding[0.5]{108}\quad Specific state-of-art use-cases/applications.  \\ \hline

 \cite{Han_Siebert_2022}& 
 
    \ding[0.5]{108}\quad  Systematic review of TinyML research.

    \ding[0.5]{108}\quad   Relevant TinyML literature on hardware, framework, data sets, use cases, and algorithms/models. 

    \ding[0.5]{108}\quad  Roadmap to understand literature on TinyML. &
    \ding[0.5]{108}\quad Existing challenges concerning multiple constraints, and future directions for research on TinyML.  \\ \hline
 
 \cite{Tsoukas_Boumpa_Giannakas_Kakarountas_2022}&
 
    \ding[0.5]{108}\quad Review of the contribution of TinyML in healthcare applications at the edge.

    \ding[0.5]{108}\quad Requirement of integration of ML followed by generated solutions. 

    \ding[0.5]{108}\quad Optimization of Neural Networks by TinyML. &
    \ding[0.5]{108}\quad Use-case related to healthcare. \\ \hline
 
    \cite{Banbury_Reddi_Lam_Fu_Fazel_Holleman_Huang_Hurtado_Kanter_Lokhmotov_etal_2020}&
 
    \ding[0.5]{108}\quad  Current landscape of TinyML. 

    \ding[0.5]{108}\quad  Challenges and directions to develop fair and useful hardware benchmark for TinyML workloads.

    \ding[0.5]{108}\quad  Benchmarks and selection methodology. &
    \ding[0.5]{108}\quad Possible solutions to future research challenges. 
 \\ \hline

\textbf{Our Survey}
&    \ding[0.5]{108}\quad Supporting software/library and platforms and associated targeted applications.

    \ding[0.5]{108}\quad Key existing challenges with respect to different constraints and possible directions. & \qquad\qquad \qquad \textbf{$-$}
 \\ \hline
    \end{tabular}}
\end{table*}

\subsection{Article Motivation}


The research on TinyML is in the early stages and requires appropriate alignments to accommodate the existing edge IoT frameworks  \cite{ray_2022}. Although initial research has demonstrated that TinyML is key to the development of smart IoT applications, several questions remain unanswered and solutions to these timely issues are required. The key issues which require immediate attention include:
\begin{itemize}
\item Key requirements and applications of TinyML.
\item Capability of TinyML to implement DNNs at the edge.
\item Power consumption versus accuracy trade-offs appropriate for and attainable by TinyML.
\end{itemize}
However, continued progress has been limited by lack of widely accepted benchmarks for TinyML-enabled systems. Specifically, bench-marking will allow measuring and thereby systematically comparing, evaluating, and improving the system's performance. This is essential to the progress of TinyML research. All in all, there are still multiple gaps in TinyML research, requiring timely solutions. This survey article lists several open research questions, outlines the potential obstacles in research on TinyML, and suggests possible directions for efficient solutions.


\subsection{Our Contribution}

Although an extensive body of literature has discussed TinyML aspects, aspects related to supporting tools and applications are not often seriously addressed. In this article, we detail the main applications  motivating TinyML research and list the corresponding supporting tools. We then detail the key TinyML enablers and advances while performing a state-of-art survey. 
Lastly, several relevant research challenges are presented followed by corresponding research directions. 

The main contributions of this article are as follows:
\begin{itemize}
\item  Presenting an intuitive understanding of TinyML and providing detailed insights regarding the related fundamentals.
\item Detailing the existing TinyML technology supporting tool-sets that are used for training the model to be deployed at edge. 
\item Discussing the key TinyML enablers and multiple use-cases/applications of TinyML. 
\item Presenting and discussing various current and future challenges in research, and related practical solutions with an outlook towards furthering research on TinyML.
\end{itemize}

Table \ref{tab:summary_of_surveys} compares the contribution of this survey article with respect to the most recent articles on TinyML. 

\subsection{Article Outline}
The rest of the paper is structured as follows. \textbf{Section II} overviews the TinyML technology and details the various tools that support TinyML. enablers. In \textbf{Section III}, we detail the multiple state-of-art applications of TinyML enabled by advanced technologies. \textbf{Section IV} identifies the various challenges and also proposes future research directions. Finally, \textbf{Section V} concludes the survey. 

\begin{figure*}[ht]
  \centering
  \includegraphics[width=15cm]{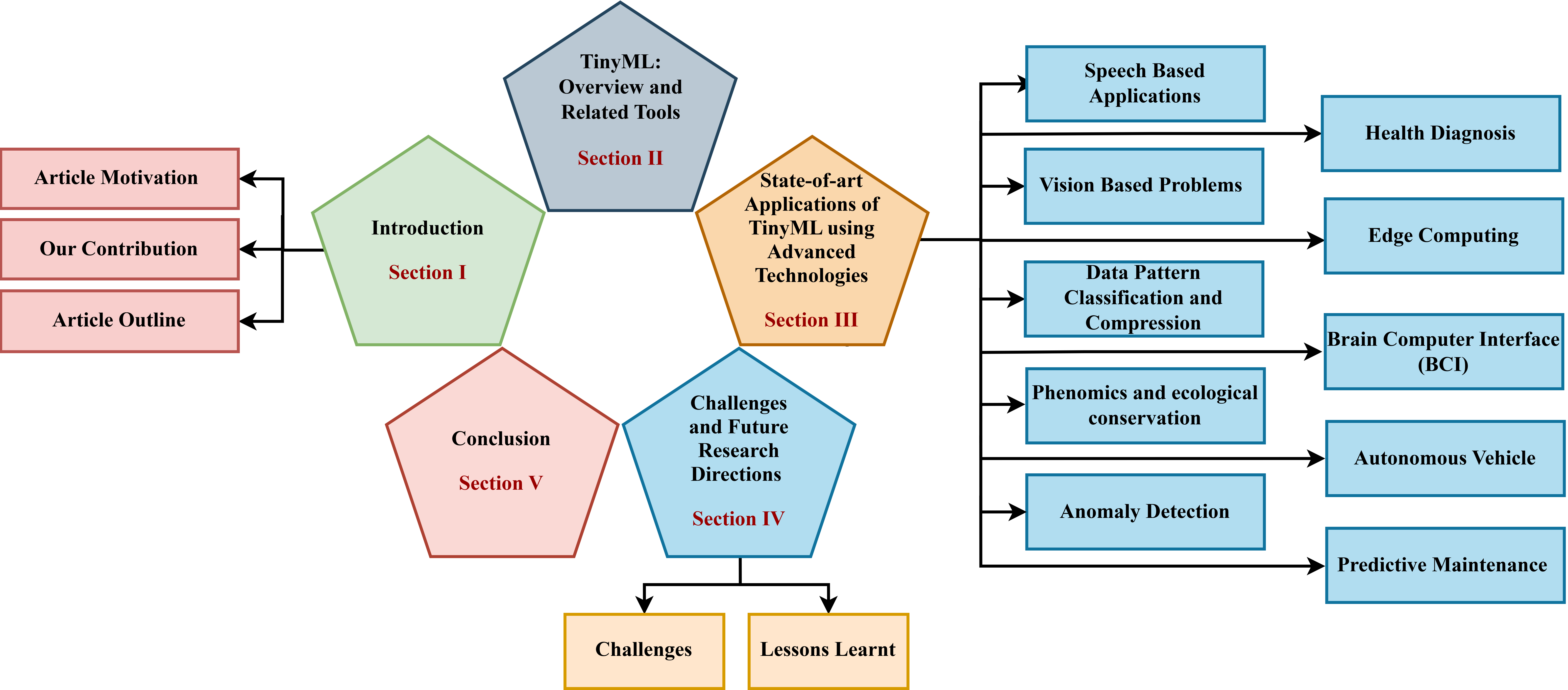}
  \caption{Taxonomy of the survey in this article.}
    \label{tab:Taxo}
\end{figure*}

Figure. \ref{tab:Taxo} shows the devised taxonomy which represents the survey on TinyML presented in this article. 

\section{TinyML: Overview and Related Tools}
TinyML can be seen as an ML tool/technique with the capability to perform on-device analytic(s) for multiple sensing modalities such as vision, audio, and speech. TinyML incurs very low power/energy consumption, thus suitable
for embedded edge devices that are battery operated. Further, TinyML is appropriate to be implemented for large-scale applications within the IoT network framework \cite{TinyMLercisson,Johnny_Knutsson_Arm_2021}.

Currently, cloud-enabled ML systems suffer a number of difficulties, including high power consumption and security, privacy, dependability, and latency issues. As a result, pre-installed models on hardware-software platforms are currently implemented (e.g., edge impulse) \cite{Gousev}. Raw data, which simulates the physical world, is gathered by sensors and subsequently processed at a CPU/microprocessor unit (MPU). The MPU aids in catering to the ML-aware analytic support enabled by specific edge aware ML networks. Notice that edge ML communicates with any remote cloud ML for transfer of knowledge. The incorporation of TinyML into system will make the physical world significantly smarter compared to any current scenario \cite{Jain}. Indeed, such a system can help edge devices to undertake key decisions even without assistance from edge AI or cloud AI. Notably, the system performance may improve over various fronts such as energy efficiency, effective data privacy, and delay. 

All in all, TinyML is envisioned as an amalgamation of hardware, software, and algorithms. Concerning hardware, IoT devices, which may or may not comprise hardware accelerators, may require analog and memory computing to provide an effective learning experience. Regarding software, TinyML applications can be implemented over cloud-enabled software or on varieties of platforms such as Linux/embedded Linux, etc. Lastly, Tiny ML systems must be supported by new algorithms requiring exceptionally low memory-sized models to avoid excessive memory consumption. Overall, the TinyML systems must optimize ML with a compact design of software in the presence of high-quality data. Then, this data must be flashed via binary files generated through the models which have been trained over a much larger machine \cite{Turnquist_Dockter_Boon_Logic_2020, Xu_Eta_Compute_2020}.


Additionally, compact software is required for small power consumption supporting TinyML implementation. Hence, systems enabled by TinyML must operate under rigid constraints while still providing high accuracy. In many cases, TinyML may rely on  energy harvesting at the edge devices to support its operation and/or enable battery-operated embedded edge devices. TinyML's fundamental requirements include (a) providing scalability to billions of cheap embedded devices, and (b) storing codes within a limited few KBs over on device RAM \cite{Gousev, Krstulovic}. Lastly, it has been demonstrated that TinyML can also be used within a standard pipeline over edge which could be changed when required by  cross-section data \cite{Alexander_Eroma_2020}. 


TinyML frameworks are continuously under development by multiple industries, specific developers, and research groups. The study conducted by \cite{9166461} shows that many of the architectures are available to the public with the exception of Fraunhofer IMS and Cartesiam-developed AlfES and NanoEdge AI Studio. Further, the majority of these frameworks support the ARM Cortex group whereas, others support the ESP8266, ESP32 group hardware, and few support Arduino and Raspberry Pi. It is evident that C and C++ are the most used languages via these architectures whereas, multiple external libraries such as TensorFlow Lie, TensorFlow, etc., can be used with the frameworks.

Several frameworks are currently being introduced by multiple research groups globally in view of implementing the TinyML models over resource-constrained devices. The study by \cite{tabanelli2022dnn} focuses on the deployment of TinyML models over edge devices to reduce the delay and improve privacy. The authors have presented a parallel ultra low power (PULP) architecture for implementation in IoT-enabled processors. PULP permits the running of non-neural ML kernels which demonstrates higher accuracy in comparison to the neural networks. PULP is compared to the PULP-OPEN hardware and it is demonstrated that PULP is 12.87x faster in comparison to ARM Cortex-M4 MCU.

An open-source toolkit namely, fast artificial neural network (FANN)-on-MCU, which runs on PULP, is developed for reducing energy consumption by \cite{9016202}. The proposed toolkit is enabled via FANN library to run the architecture, implemented in the InfiniWolf prototype, over lightweight IoT-aware devices. The authors have compared FANN-on-MCU with RISC-V octa-core processor and have demonstrated that FANN-on-MCU incurs an increased speedup by 22x and consumes 69 percent less energy.

In \cite{fahim2021hls4ml}, the authors have presented hls4ml TinyML architecture for reducing energy consumption. The proposed framework enables the acceleration of ML-aware FPGA and ASIC implementation in a feasible and easy manner and provides Python-based APIs to harness scientific benefits from the framework. Further, it also provides  quantization and pruning-aware training for low-power embedded devices. 

In \cite{Paissan}, PhiNets, a scalable backbone architecture for DNNs is detailed which is designed for providing image processing application support for resource-constrained edge IoT devices. The proposed framework is developed over the inverted residual blocks to decouple cost, memory, and over-processing. The results demonstrated that PhiNets reduces the count of parameters by 85–90-\ in comparison to existing architectures. 

In \cite{Bringmann}, to address the hardware/software co-design, the HANNAH framework is detailed which aims at automating co-optimization steps of a NN framework for efficient end-to-end DNN training and use over edge devices. HANNAH is implemented in 3 steps with the Ultra-Trail NN unit. 

Following \cite{Alajlan_Ibrahim_2022}, each industry has unique software and ML model to make the embedded systems board compatible with TinyML applications. However, with TinyML this may present significant issues. TinyML is a technique for using extremely compact computer programs for tasks such as voice recognition and motion detection. However, it will be challenging to utilize TinyML over several devices without compromising on accuracy since each industry has its own software. Hence, it will be crucial to provide a standardized approach for implementing TinyML. This necessitates the development of a common framework that can run on various hardware manufactured by various industries. Once successful, this will ensure that multiple commonplace applications can be provisioned by TinyML.

From the above, it can be inferred that among the multiple resource constraints viz., data set generation, execution time, etc., hardware platforms present the key constraints for TinyML's high performance. Hence, there is a need to encourage the design and training of any TinyML model using specific software/library/framework and deploy the trained model to the supporting hardware platform.  

In Table \ref{tab:related}, we list the details of the available hardware platforms supporting the design environment i.e., frameworks/libraries. Further, we also list the software/libraries which can be integrated with the related hardware platforms to provision specific application(s)/use-case(s). Our survey reveals that approximately (i) 59 \% of participants use TensorFlow, (ii) 17 \% of participants use PyTorch, and (iii) 24 \% of participants use other frameworks. Further, the top 3 data types used by ML users are vision data, motion data, and sound data. In regard to the hardware boards, the most used for developing TinyML projects include (i) Raspberry Pi, (ii) Arduino Nano 33 BLE Sense, (iii) ESP32, and (iv) Raspberry Pi Pico and NVIDIA Jetson Nano. 



\begin{table*}[ht!]
 \caption{Supporting platforms for TinyML.} 
     \label{tab:related}
    \centering
    \resizebox{18cm}{!}{
    \begin{tabular}{|p{1.5cm}|p{2cm}|p{2cm}|p{2cm}|p{4cm}|p{4cm}|}
\hline
  \textbf{Reference} & \textbf{Software/ Library / Framework} & \textbf{Developer} & \textbf{Supporting Platform} & \textbf{Hardware Platforms} & \textbf{Targeted Applications}\\
  \hline 

\cite{tensorflowinfer}& TensorFlow Lite (TFL) & Google Brain Team &
Android, iOS, Embedded Linux, Micro-controllers &
Arduino Nano 33 BLESense, Sparkfun Edge, STM32F746 Discovery Kit , Adafruit Edgebadge, Adafruit TensorFlow Lite for Microcontrollers Kit, Adafruit Circuit Playground Bluefruit, Espressif ESP32-Devkitc, Espressif ESP-EYE, Wio Terminal: ATSAMD51, Himax WE-I Plus EVB Endpoint AI Development Board, Synopsys Designware ARC EM Software Development Platform, Sony Spresense&
Image and Audio Classification, Object Detection, Pose Estimation, Speech and Gesture Recognition, Segmentation, Video Classification, Text Classification, Reinforcement Learning, On Device Training, Optical Character Recognition \\  \hline

\cite{microTensor} & Utensor & ARM &
Android, iOS, Embedded Linux, Micro-controllers &
Mbed, ST K64 ARM Boards &	
Image Classification, Gesture Recognition, Acoustic Detection and Motion Analysis \\ \hline

\cite{edge} &Edge Impulse	&Zach Shelby And Jan Jongboom &
Android, iOS, Embedded Linux, Micro-controllers &
Arduino Nano 33 BLE Sense, Arduino Nicla Sense ME, Arduino Nicla Vision, Arduino Portenta H7 + Vision Shield, Espressif ESP32, Himax WE-I Plus, Nordic Semi Nrf52840 DK, Nordic Semi Nrf5340 DK	&
Asset Tracking and Monitoring, Human Interfaces, Predictive Maintenance \\ \hline

\cite{nano} & Nanoedge AI Studio &	Cartesiam &
Android, Linux &
STM32 Boards &
Anomaly Detection, Predictive Maintenance, Condition Monitoring, Asset Tracking, People Counting, Activity Recognition\\ \hline

\cite{pytorch} & Pytorch Mobile &	Meta AI (Facebook)&	
Android, iOS, Linux	CPU &
NNAPI (Android), Coreml (iOS), Metal GPU (iOS), Vulkan (Android)&
Computer Vision and Natural Language Processing \\ \hline

\cite{ell} & Embedded Learning Library (ELL)	& Microsoft	&
Windows, Ubuntu Linux, Mac OS X &
Raspberry Pi, Arduino, Micro: Bit	&
Image And Audio Classification \\ \hline

\cite{stm} & STM32Cube.AI	&STMicroelectronics &
Android, Linux&
STM32 ARM CORTEX Boards	&
Anomaly Detection, Predictive Maintenance, Condition Monitoring, Asset Tracking
People Counting, Activity Recognition \\ \hline

\cite{Sun_Vlasic_Herrmann_Jampani_Krainin_Chang_Zabih_Freeman_Liu_2021, tinyframe, autooptical} & Autoflow &	Daniel Konegen And Marcus Rüb	&
Android, iOS, Embedded Linux, Micro-controllers, &
MCU, FPGA Boards, Raspberry Pi	&
Image Classification, Object Detection, Pose Estimation, Speech Recognition, Gesture Recognition \\ \hline

\cite{apache} & Apache Mxnet &	Apache Software Foundation (ASF)&
Linux&
Raspberry Pi, NVIDIA Jetson	&
Image Classification, Object Detection, Pose Estimation, Speech and Gesture Recognition \\ \hline

\cite{mlkit}& ML Kit for Firebase &	Google &	
Android, iOS &
Mobile Devices	&
Facial Detection, Bar-Code Scanning, Object Detection \\ \hline
    \end{tabular}}
\end{table*}

\section{State-of-art Applications of TinyML using Advanced Technologies}

There are several applications of TinyML, including speech and vision-based applications, data pattern classification and compression, health diagnosis, edge computing, brain-control interface, autonomous vehicles, phenomics, and ecology monitoring. This section details the state-of-art applications of TinyML using various advanced technologies.

\subsection{Speech-Based Applications}
\subsubsection{Speech Communications}
\begin{figure}[ht]
  \centering
  \includegraphics[width=9cm]{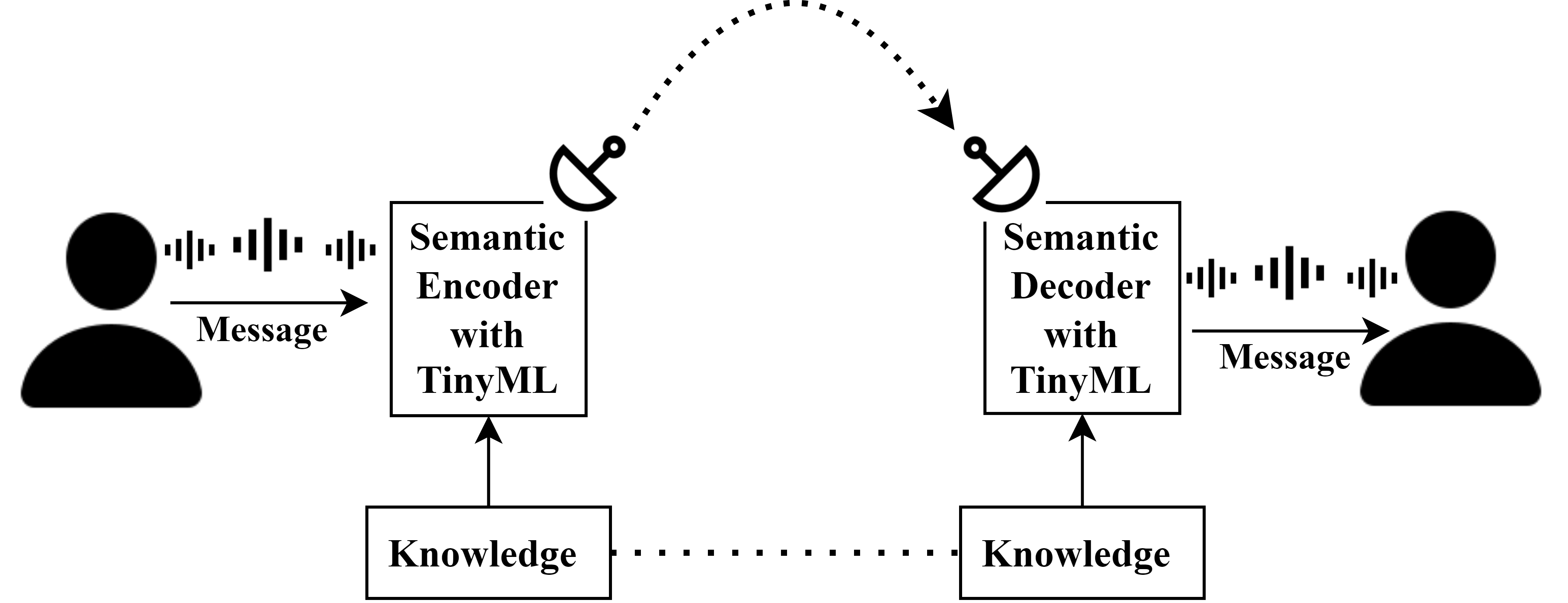}
  \caption{Speech recognition and response.}
    \label{tab:speech}
\end{figure}

Semantic communication emerged as an alternative to conventional communication. In the latter, all the data matters and is transmitted, while in the case of the former, only the context/meaning of the data is transmitted to the receiver. Notably, semantic communication can be implemented by employing the TinyML methodologies \cite{Iyer_Khanai_Torse_Pandya_Rabie_Pai_Khan_Fadlullah_2022}. Speech detection and recognition, online teaching/learning, and goal-oriented communication as shown in Figure \ref{tab:speech}, are popular applications in the current scenario and require high data and high-power consumption on the host device. To overcome these drawbacks, TinySpeech library has been introduced to build a low computational architecture with a low storage facility using deep convolutional networks \cite{Alajlan_Ibrahim_2022}.

In view of Speech Enhancement \cite{Fedorov_Stamenovic_Jensen_Yang_Mandell_Gan_Mattina_Whatmough_2020}, the authors addressed sizing of the speech enhancement model as it was subjected to hardware resource constraints. The study employed structured pruning and integer quantization for the Recurrent Neural Network (RNN) speech enhancement model. The results suggested reducing the model size by 11.9x and operations by 2.9x. The study also demonstrated the usefulness of hearing aid products enabled by neural speech enhancement methods with better battery life. 
Resources must be utilized efficiently for energy-constrained edge devices executing voice-recognition applications, as demonstrated by the authors in \cite{Kwon_Park_2021}. Therefore, the study aimed to partition the process and proposed a co-design for the TinyML based voice-recognition. The authors used windowing operation to partition hardware and software such that raw voice data would be pre-processed. The results demonstrated a decrease in energy consumption on the hardware. Lastly, the authors proposed optimized partitioning among hardware and software co-design as future scope.

Authors in \cite{Zhang_Sun_Ma_2021} proposed a phone-based transducer for the speech recognition system. However, the study involved replacing the LSTM predictor with the conv1d layer for reducing the computations on the edge device. The results revealed that the Singular Value Decomposition (SVD) technology had successfully compressed the model. While the Weighted Finite State Transducers (WFST) based decoding allowed the flexible bias in the model improvement. 
A similar study on speech recognition systems was conducted in \cite{Li_Alvarez_2021} by introducing integer quantization in the LSTM neural network topology to reduce memory consumption and computation latency. The results demonstrated that the proposed model achieves good accuracy even on a data set that consists of long utterances.

Live captioning, virtual assistants, and voice commands are all prominent applications of SR and all of them require ML for their work. The current SR technologies (such as Siri and Google Assistant) have to ping the cloud every time they receive data, which creates concerns regarding data security. The solution for this problem is to perform on-device SR, which is where TinyML comes into play. Authors in \cite{zhang2021tiny} proposed Tiny Transducer, an SR model for the on-device scenario, which uses a deep feed-forward sequential memory block (DFSMN) layer on the encoder side and one Conv1d layer on the predictor side in place of LSTM layers to bring down both network parameters and computation.

\subsubsection{Hearing Aid (HA)}
It is common that most people in their later half experience hearing loss. This health issue constitutes a serious problem for countries dealing with population aging, e.g., Japan, South Korea, and China, opening a business opportunity for the HA industry. However, a critical problem must still be resolved and it is that currently, HA devices amplify all of the input sounds, thus making it difficult for a person to distinguish the desired sound in a noisy environment. According to \cite{fedorov2020tinylstms}, TinyML can provide a solution to this problem. Therein, the authors  proposed a TinyLSTM-based speech enhancement (SE) algorithm for HA devices, which performs the denoising operation over the input sounds and extracts the speech signal. When tested on STM32F746VE MCU and trained with the CHiME2 WSJ0 data-set \cite{6637622}, the algorithm shows a computational latency of $2.39 ms$, which is far less than the $10 ms$ target.

\subsection{Vision-Based Applications}
To process computer vision based data-sets, TinyML can play a crucial role as these processes need to be performed on the edge platform to generate faster outputs. Authors in \cite{Paul_Mohan_Sehgal_2020} addressed the practical challenges in training the model using the OpenMV H7 micro-controller board. They proposed an architecture for detecting alphabets within American Sign Language over ARM Cortex-M7 microcontroller with only 496 KB of frame-buffer RAM. In effect, the authors addressed the major challenge of convolutional networks (CNNs) with high generalization error, including large test and training  accuracy. However, these did not generalize effectively to images within new cases and backgrounds with noise. The authors employed interpolation augmentation; results show 98.80\% accuracy in test and 74.59\% accuracy in generalization. It was observed that interpolation augmentation reduced drop in accuracy during quantization in hand sign classification. However, it improved classification generalization (with a 185 KB post quantization model) and inference speed (to 20 fps). The authors also proposed the future scope to improve accuracy in generalization model training on data from highly varied sources and testing it over hardware to attain the ambition of portable watch-like device. 
We know that word level vocabulary and non manual features require identification of facial expressions, mouth, tongue, and body pose. By extending the study on CNN, the authors in \cite{Mohan_Paul_Chirania_2021} deployed CNN architecture on a resource-constrained device. They developed framework to detect medical face masks over resource constrained ARM Cortex M7 micro-controller using TensorFlow lite with extremely low memory footprints. The results demonstrated 138 KB model size post quantization and 30 frames per second inference speed on the targeted board. The authors also proposed the research scope on developing (i) quantization schemes for reducing the precision from float32 to int8, (ii) data sets with heterogeneous sources, and (iii) experiments with smaller precision networks.

Authors in \cite{Patil_Dennis_Pabbaraju_Shaheer_Simhadri_Seshadri_Varma_Jain_2019} presented a case study aiming to design a gesture recognition device that could be clamped to an existing cane to be used by the visually impaired. The design constraints considered were low cost, accurate gesture detection, and battery design. Further, the data was collected using a gestures data set, and the ProtoNN model was trained with a classification algorithm. Finally, as a scope for future research, the authors mentioned the necessity of understanding gestures and their associated safety and the integration of android and developed devices. In \cite{dePrado_Rusci_Capotondi_Donze_Benini_Pazos_2021}, the authors addressed challenges, such as resource scarcity and on-board computation, faced in scaling the autonomous driving to mini-vehicles. The authors introduced a TinyCNN-based closed-loop learning flow and proposed an online predictor model which takes into account the recently captured image at the run-time. The major challenge observed in the design of autonomous driving was the decision model developed for offline data, which may not be robust for online data. For such applications, the authors stated that the model design should be able to adapt to real-time data, and this motivated the authors' current study. The authors performed experiments on GAP8, STM32L4, and NXP k64f. It was demonstrated through comparative results that GAP8 outperforms in terms of energy consumption and latency with online data. As a future study, CNN on-chip can be trained for continuous learning considering real-time applications.

The study on NAS was conducted by authors in \cite{Benmeziane_Maghraoui_Ouarnoughi_Niar_Wistuba_Wang_2021}, where NAS was implemented in image classification and object detection problems. The authors addressed the real-time challenges such as deploying architectures of synthesized CNN, and the hardware-aware NAS was proposed as a solution. In addition, the study involved a detailed survey of the challenges and limitations of existing approaches, and their categorization with respect to acceleration techniques, cost estimation of hardware, search space, and search strategy was also available.

\subsection{Data Pattern Classification and Compression}
The challenge of adapting a trained TinyML model to the online data has attracted attention from the research community. The authors in \cite{Ren_Anicic_Runkler_2021} proposed a novel system, namely, TinyML with Online Learning (TinyOL), for introducing training with incremental online learning on MCU and enabling updating the model online on edge devices of IoT. The implementation was performed using C++ language, and an additional layer was added to the TinyOL. Further, the study was performed on the auto-encoder of Arduino Nano 33 BLE sense board \cite{tensorflowinfer}, and the model was trained to classify new data patterns. The research scope mentioned included the design of efficient and optimized algorithms for the neural network to support online device training patterns.
Authors in \cite{Cai_Gan_Zhu_Han_2020} mentioned the number of activation layers as a major issue for memory-constrained AI edge devices. As a result, Tiny-Transfer-Learning (TinyTL) was introduced to efficiently utilize the memory over an edge device and avoid using the intermediate layers as activation. In addition, to up-hold the adaptation capabilities and allow the feature extractor to discover the small residual feature maps, a bias module known as the 'lite residual module' was also introduced. Compared to the full network fine-tuning, the results showed TinyTL reduced the memory overhead by 6.5x with a moderate accuracy loss. While compared to the case when the last layer was fine-tuned, TinyTL displayed a 34.1\% of accuracy improvement once again with a moderate accuracy loss.

Authors conducted a detailed study on data compression in \cite{Signoretti_Silva_Andrade_Silva_Sisinni_Ferrari_2021}, emphasizing that data compression algorithms must manage extensive collected data in a portable device. The authors developed Tiny Anomaly Compressor (TAC) and demonstrated that TAC outperforms Swing Door Trending (SDT) and Discrete Cosine Transform (DCT) algorithms. Furthermore, TAC achieved a maximum compression rate of 98.33\% and outperformed both SDT and DCT in terms of peak signal-to-noise ratio.

\subsection{Health Diagnosis}
With the spread of COVID-19, it is now required to continuously detect cough-related respiratory symptoms. The authors in \cite{Rashid} have presented a scalable CNN enabled model namely, Tiny RespNet which operates over multi-modal settings. These settings are deployed over Xilinx Artix-7 100 t FPGA which provides the parallel processing facility with low power consumption and high energy efficiency. Further, Tiny RespNet framework is able to input audio recordings, speech of patient, and information of demography to ensure classification. The cough detection related respiratory symptoms are classified via three data sets.

The authors in \cite{Coffen_Mahmud_2021} conducted a study on deep learning computations on edge devices. The study aimed to implement a TinyML model named TinyDL on wearable devices for health diagnosis and carried out the performance accuracy analysis to reduce latency, bandwidth, and power consumption. A multi-layer LSTM model was designed and trained for a wearable device with the collected accelerated data as the input. The model exhibited an accuracy of 75-95\% accuracy per gesture and was able to analyze only the off-device data. This limitation was due to compatibility issues of the framework, which authors corrected in \cite{Alajlan_Ibrahim_2022} by providing a detailed study of TinyML's significance, and related role in IoT. The ML models on edge devices showed the potential solution to the existing challenges of IoT. The authors also provided a detailed review of models, device types, and the data sets used in TinyML, in addition to a detailed survey of the existing and supporting framework on platforms.

Another application includes the estimation of body pose which is vital to monitor health of the elderly. In \cite{vuletic2021edge}, platform agnostic framework is proposed to enable validation and rapid fostering of model to platform adaptations. It implements face landmarks and body pose estimation algorithms and implements composite fields to detect spatiotemporal body pose in real time. The platform used is Nvidia Jetson NX consisting of GPUs and DL hardware accelerators.

\subsection{Edge Computing}
\begin{figure}[ht]
  \centering
  \includegraphics[width=9cm]{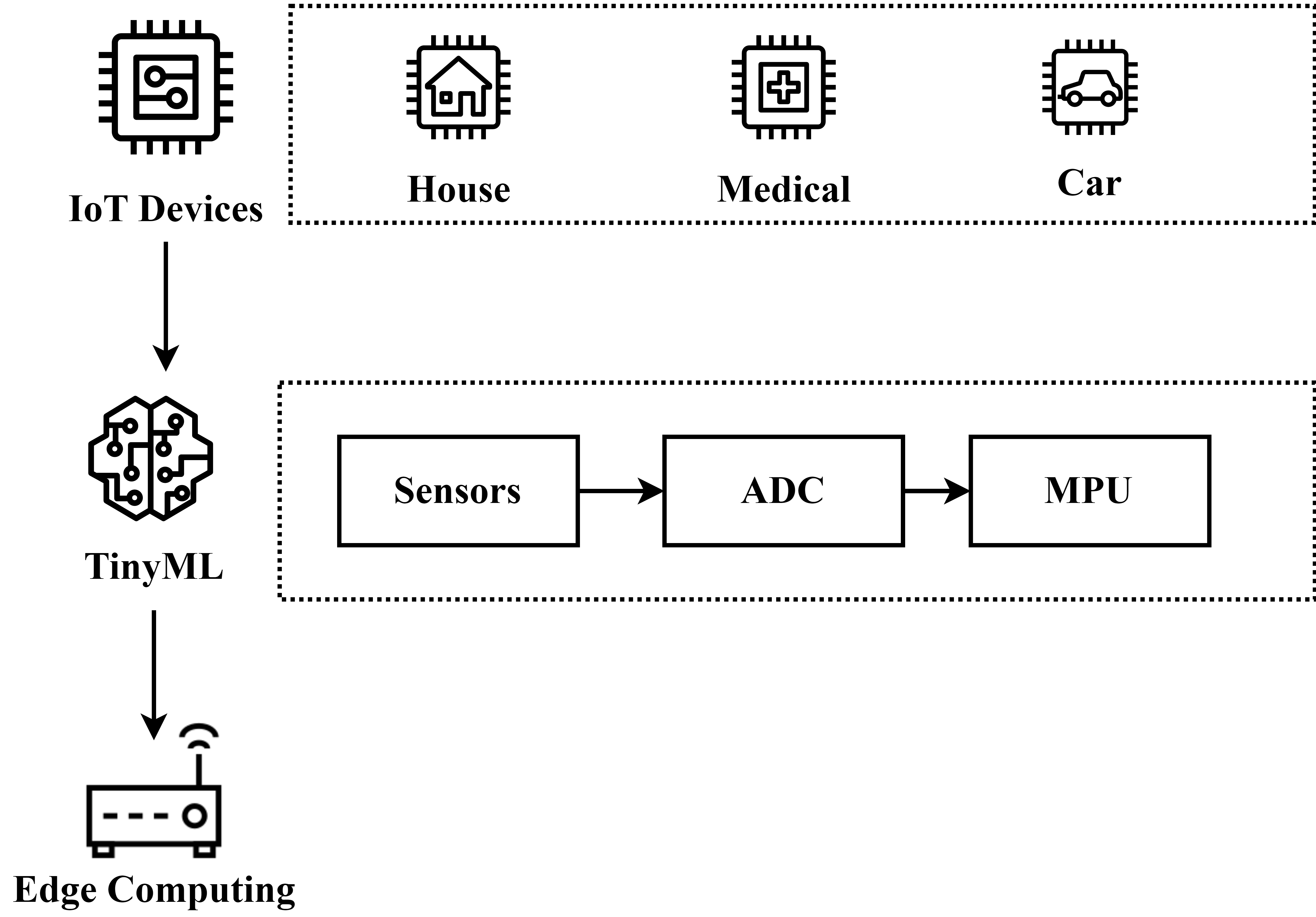}
  \caption{Edge devices combined with TinyML for Edge Computing.}
    \label{tab:edge}
\end{figure}

With the massive increase in IoT devices connecting to the global network, there is an urgent need for setting up edge devices to reduce the load on the cloud. These edge devices carry individual data centres capable of high-level computing, which  results in high security, and reduced cloud dependency,  latency, and bandwidth. The edge devices enriched with TinyML algorithms will help in fulfilling the power, memory, and computing time constraints as shown in Figure \ref{tab:edge}. In \cite{Raza_Osman_Ferrini_Natale_2021}, the authors detailed the energy efficiency problems faced during the practical implementation of an edge  Unmanned Aerial Vehicle (UAV) device. The goal was to implement an energy-efficient device with low latency by interfacing TinyML on the MCU, which acts as a host controller for the UAV. For various edge computing activities, there is a need for sensors for data acquisition. Edge sensors such as blood pressure sensors, accelerometers, glucose sensors, Electrocardiogram (ECG) sensors, motion sensors, and Electroencephalogram (EEG) sensors are widely used for the data gathering process during edge computing \cite{Awad_Fouda_Khashaba_Mohamed_Hosny_2022}.

\subsection{Brain-Computer Interface (BCI)}
\begin{figure}[ht]
  \centering
  \includegraphics[width=9cm]{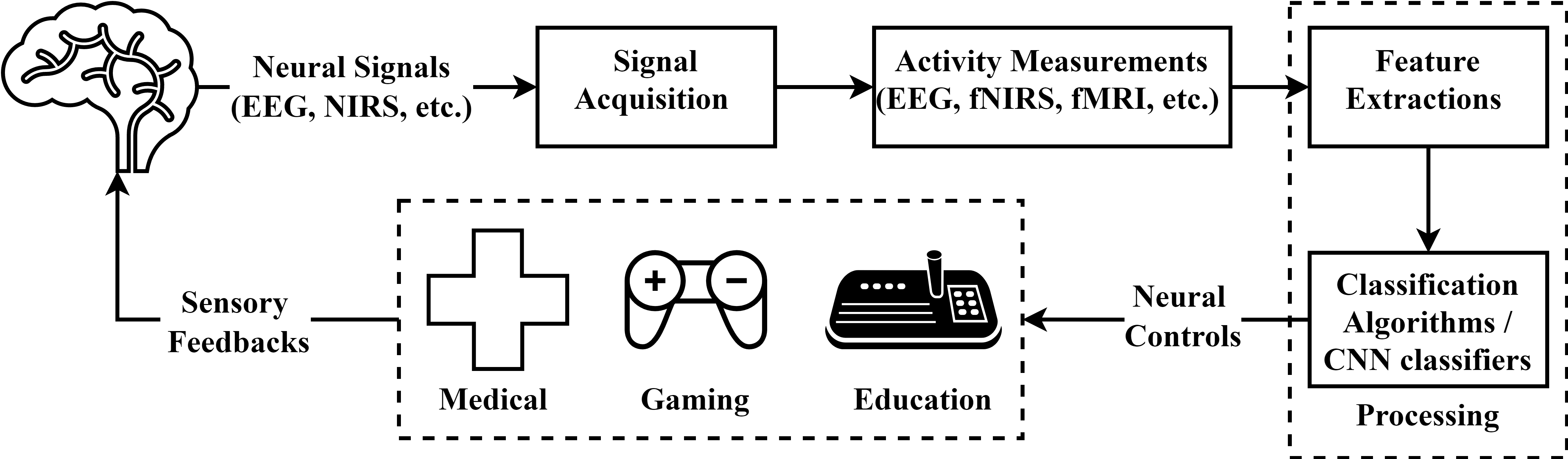}
  \caption{BCI Methodology.}
    \label{tab:bci}
\end{figure}

Within the healthcare sector, TinyML can contribute significantly; to, e.g., tumor and cancer detection, emotional intelligence, and health predictions using EEG and ECG signals \cite{Pai_Kallimani_Iyer_Maheswari_Khanai_Torse_2022} as shown in Figure \ref{tab:bci}. With the aid of TinyML technology, Adaptive Deep Brain Stimulation (aDBS) \cite{Merk_Peterson_Kohler_Haufe_Richardson_Neumann_2022} has the potential to demonstrate breakthroughs in successful clinical adaptations. aDBS helps in the identification of disease-specific bio marks and their respective symptoms through invasive recordings of the brain signals. Further, as healthcare mainly includes a collection of enormous data and then processing it to reach specific solutions for the early cure of the patient, it is necessary to build a system that is extremely accurate and highly secure. Such a system in the medical science field, when combined with IoT and TinyML, is termed as the Healthcare Internet of Things (H-IoT) \cite{Bharadwaj_Agarwal_Chamola_Lakkaniga_Hassija_Guizani_Sikdar_2021}. The major applications of H-IoT are monitoring, diagnosis, spread control, logistics, and assistive systems. To detect a patient's health state remotely there is a need to develop a highly reliable system with extremely low latency and global accessibility. This system can be developed by integrating H-IoT with TinyML and 6G-enabled Internet services \cite{Padhi_Charrua-Santos_2021}.

\subsection{Autonomous Vehicle}
Autonomous vehicles are utilized during multiple emergency cases such as military, human tracking, and industrial applications. Such vehicles demand smart navigation to ensure efficient identification of the object(s) being searched. Currently, autonomous driving is a complex task especially, when lower scale such as mini-vehicles are desired. The TinyML technology can be implemented to improve autonomous driving of the mini vehicles as shown in \cite{Prado} where it has been deployed over the GAP8 MCI which is enabled by the framework of convolution neural network (CNN). Further, the same is tested over STM32L4 and NXP k64f platforms. The results demonstrated that such an integration reduces the processing delay by 13 times and simultaneously provides an improvement in consumed energy of approximately 90\%. In addition, automatic traffic scheduling has been investigated in recent years to possibly integrate it with the TinyML technology in view of improving real time traffic system \cite{9419472}. The proposed method includes piezoelectric sensors which are embedded over the multiple lanes of a specific road, and the two point time ratio technique is used for detecting vehicle by using data from piezo-sensors. Further, vehicle classification includes prediction of green light timings. The study implements the random forest regressor for predicting signal duration depending on the count of input vehicles over every lane. The implementation is over an Arduino Uno which is supported by the m2gen library using Scikit Learn requiring only 1.754 KB for the algorithm.

\subsection{Phenomics and conservation of ecology}
Phenomics is defined as study of phenotypes related to genome wide changes in an organism’s lifespan. Specifically, among the entire germplasm set, plant phenomics utilizes improvements in genes to discriminate key germplasm. The study by \cite{Nakhle_Harfouche_2021} performed a phenomics image analysis which is based on tomato leaf disease and spider mites classification. The method utilizes Plant-Village tomato data set in conjunction with YOLO3 algorithm which is enabled by DarkNet-53 architecture to automatically detect tomato leaves. The study further implements SegNet algorithm to segment the images in a pixel-wise manner. Lastly, the study investigates multiple other commonly used data analysis tools for validating phenomics use with TinyML. In recent years, ecological conservation analytics enabled by AI techniques has witnessed tremendous growth. The study by \cite{Curnick_2022} has deployed ecology conservation step using TinyML within small payload satellites (SmallSats). Specifically, study focuses on improvement of sea turtles’ conservation using advanced real time vision based TinyML. Another crucial application of TinyML includes environment monitoring. In \cite{9239623}, a deep tiny NN is detailed for prediction of weather. The proposed framework utilizes STM32 MCU and X-CUBE-AI tool chain over the Miosix operating system. The framework requires 45.5 KB of flash and 480 Bytes onboard RAM for running deep tiny NN architecture.

\subsection{Anomaly detection}
Anomaly is defined as an event that varies from majority mass of events. In \cite{lord2021tinyml}, an investigation is conducted for finding appropriateness of TinyML to detect anomalies of related tasks. A generic ANN, auto-encoder, and variational auto-encoder are used over Arduino Nano 33 BLE sense module, and top load washing machine Kenmore model is implemented to detect anomalies over unbalanced spin dry cycle. The results demonstrated approximately 90\% precision and accuracy.

\subsection{Predictive Maintenance} 
Predictive Maintenance has emerged as a promising maintenance paradigm in which models perform the prediction of equipment failure \cite{primer}. Currently, the majority of the predictive maintenance systems have been used either over cloud or via powerful computers. The data utilized by such models are mostly generated by tiny sensor devices and hence, the current approach requires data to be aggregated and transmitted over network for processing. TinyML can be used to implement an alternative to cloud-based predictive maintenance systems. This will require the optimization of the entire TinyML pipeline in an industrial setting which if achieved, will demonstrate immense potential for input optimization in view of achieving predictive maintenance using TinyML.

\begin{table*}[ht!]
 \caption{Summary of multiple existing challenges on TinyML research and suggested directions to obtain solutions \cite{ray_2022}.} 
     \label{tab:summary}
    \centering
    \resizebox{18cm}{!}{
    \begin{tabular}{|p{1.5cm}|p{3cm}|p{6cm}|p{6cm}|}
\hline
  \textbf{Sl. No.} &\textbf{Constraint} &\textbf{Existing Challenges} &\textbf{Proposed Directions}\\
  \hline 
1.		&Resource	& Limited power availability at the edge devices is a critical challenge to maintaining the algorithm’s accuracy.	&Design the edge devices with a co-design approach to meet the power management challenges. \\
\cline{3-3} \cline{4-4}
	&	& Limited memory size is another challenge as deploying the model needs a higher peak of memory.	&Quantise the model and use an appropriate converter to convert from float type to integer type and use the memory on edge hardware appropriately.\\ \hline
	
2.		&Hardware	&Existing benchmarks need to be re-framed before deploying TinyML on the resource constraint hardware	&Redesign the benchmark to balance the resource constraint and data heterogeneity data of the system. \\ \cline{3-3} \cline{4-4}
	&
		&Heterogeneity is a challenge as there are various embedded hardware available over a wide area of application; an extremely heterogeneous ecosystem demands a different type of micro-controllers and the model generated is expected to work efficiently on the targeted embedded hardware	& Develop a generalized model so that it can work efficiently with heterogeneous systems \\ \hline

3.		&Data-set	&The architecture of TinyML systems do not support the existing data set. This is a challenge to all edge devices as the data collected from external sensors need resolution and the devices are energy and power constrained. Thus the existing data set cannot be used directly to train the TinyML models.   	& Train the TinyML model with a standard and optimized data set.
\\ \cline{3-3} \cline{4-4}
	&
		&The lack of popularly accepted Models is another challenge in the research domain	&Develop a model which can be readily adopted by the system and improve the TinyML ecosystem
\\ \cline{3-3} \cline{4-4}
	&
		&The heterogeneous data type is a major concern, especially for data and Network Management	&An intelligent network is needed as current data and networks are not managed for the data.\\ \hline
4.		&Existing edge infrastructure	&Edge computing infrastructure faces a challenge as the resources are changing dynamically, in turn affecting the TinyML ecosystem.	&The need for enablers is key to providing support to system and leveraging the existing infrastructure\\ \cline{3-3} \cline{4-4}
	&
		& Currently, the edge platform suffers from the issue with dynamic resource allocation due to which there is a need for techniques/ algorithms for the analysis of dynamic data.	&Employing optimization techniques could be one of the directions to support dynamic edge resource allocation. \\ \hline
5.		&Design of ML Models	& The response time of ML models is another issue discussed in the research community widely.	&Models need to be co-designed to provide good and quick responses for all the edge devices with model pruning and quantization being a part of model design. \\ \hline

    \end{tabular}}
\end{table*}

\section{Challenges and Future Research Directions}

In this section, we present the various challenges and issues in research related to the multiple TinyML applications. Specifically, we summarise key leanings on TinyML from the detailed discussions throughout the paper. Next, we have also detailed the possible solutions to provide future scope to the researchers for contributing towards providing solutions to the various issues and challenges. 
 
\subsection{Challenges}

TinyML encounters major hurdles hindering the related growth pattern. The key challenges include the following:
\begin{itemize}
 \item Currently, for the embedded edge IoT devices battery power consumption is expected to be over a period of ten years \cite{Media}. For e.g., in ideal conditions, a battery capacity of 2Ah is expected to have a life cycle of more than ten years considering that the power consumption is less than 12µA. However, when a simple edge IoT device's circuit is considered with a combination of MCU, temperature sensor, and a Wi-Fi Module, the aggregate current consumption is approximately 176.4 mA \cite{Consumption_2022}. This will invariably reduce the life cycle of the 2Ah battery to approximately 11 hours and 20 minutes. This aspect is a critical challenge in the TinyML ecosystem.
 \item Majority of edge devices operate at clock speeds between 10–1000 MHz which is restrictive in the effective execution of complex learning models at the edge \cite{Gousev}. 
 \item Memory is limited. Indeed, existing TinyML edge platforms operate with lesser than 1 MB onboard flash memory \cite{Gousev}. This restricts the performance of models and presents a significant challenge in view of accommodating the MCU. 
 \item The cost of large-scale deployments can be significant even when the cost per device is low. Thus, for the success of low-cost edge platforms, monetary issues must be addressed \cite{Situnayake}. 
\end{itemize}

\subsection{Lessons Learnt}
Through our comprehensive survey, we identified that the key issue in research over TinyML includes low availability of power within edge devices, hence, calling for energy-efficient TinyML system designs. In this regard, efficient energy harvesting techniques must be implemented for powering smart devices \cite{Lopez.2021}, so that an appropriate amount of energy is dedicated to ML-related tasks. Limited memory is another factor that hinders the growth of TinyML. Thus, research must be focused on the low memory footprint of edge hardware for the TinyML systems. In terms of problems related to clock speeds, much research attention must be focused on providing the optimal solution to address the issues related to the capacity of the processor. Also, running complicated ML algorithms on MCU is difficult due to the low CPU capability.

The heterogeneity in the infrastructures of hardware/software poses significant challenge to TinyML systems in view of adopting a specific learning mechanism and deployment strategy. Also, existing domain related to edge computation is in an early stage which does not allow for the adoption of resources that change dynamically within the edge devices. Hence, it will be required to include device mobility and reliability factors during the deployment of ML models. In regard to reliability, the significant issues will be in terms of variations in process, hard and soft errors, and aging. Hence, it will be important to ensure that a specific edge device undergoes the reliability assessment before it is deployed for any application.

With the implementation of TinyML, new ML models will have to be formulated for introduction within the TinyML ecosystem. Techniques such as federated learning, transfer learning, and reinforcement learning can be used for the model design which must provide real-time solutions. In regard to edge-based solutions, the edge infrastructure must be based on techniques of virtual optimization for supporting multiple level dynamics at edge. Considering software for the edge, such a design will require specialized skill sets. 
The merging of edge devices and software is an additional challenge to be addressed. 
Overall, the edge intelligence framework must provision advanced applications such as 5G/6G wireless networking, data and cooperative intelligence, management of energy efficiency, ML as a service, etc. 

Finally, the lack of bench-marking tools, data sets, and accepted models also presents a key challenge and must be resolved for TinyML research to move forward. The absence of standardization is one of the main problems due to which it is challenging for developers to generate cross-platform compatible solutions. 

Table \ref{tab:summary} lists the issues and challenges which have been identified, and also presents the possible solutions in terms of the proposed directions.  


\section{Conclusion}
The development of ML techniques has led to a paradigm shift in the IoT ecosystem. Indeed, the integration of ML at the edge devices may ensure that the IoT systems can take intelligent decisions. As these edge devices are resource constrained, there is immense interest from the research community to implement low-complexity ML techniques, i.e., TinyML. The TinyML technology allows the tiny devices to be optimized, thereby ensuring accuracy and efficiency. In this article, we have surveyed the emerging growth of TinyML in the field of edge and energy computing IoT devices. The implementation of TinyML requires training the models, performing quantization techniques, and deploying the trained model on the hardware. The ML models to be deployed on edge devices have multiple research challenges due to the involved complexities, including the selection of hardware and compatibility of the framework. This article provides a detailed study of the available hardware platforms and software frameworks to encourage the growth potential of TinyML. We have also presented future research directions to various existing challenges in view of spurring future research on TinyML.

Despite the multiple difficulties, TinyML has immense potential to revolutionize multiple sectors including manufacturing, transportation, agriculture, and healthcare. We anticipate future creative uses of the TinyML technology as further studies are conducted and advanced use cases/applications emerge. Finally, through this comprehensive survey, we hope to have extended research on key issues related to TinyML and aid successful implementations of multiple TinyML applications.


%

\ifCLASSOPTIONcaptionsoff
  \newpage
\fi



%

\bibliographystyle{IEEEtran}
\bibliography{references.bib}

%




\end{document}